# Simulation and Calibration of a Fully Bayesian Marked Multidimensional Hawkes Process with Dissimilar Decays

**Kar Wai Lim**　　　　　　　　　　　　　　　　　　　　　　　　　karwai.lim@anu.edu.au
Data61 / CSIRO
Australian National University

**Young Lee**　　　　　　　　　　　　　　　　　　　　　　　　　young.lee@data61.csiro.au
Data61 / CSIRO

**Leif Hanlen**　　　　　　　　　　　　　　　　　　　　　　　　leif.hanlen@data61.csiro.au
Data61 / CSIRO

**Hongbiao Zhao**　　　　　　　　　　　　　　　　　　　　　　　　　h.zhao1@lse.ac.uk
Department of Finance, SOE & WISE, Xiamen University

## Abstract

We propose a simulation method for multidimensional Hawkes processes based on superposition theory of point processes. This formulation allows us to design efficient simulations for Hawkes processes with differing exponentially decaying intensities. We demonstrate that inter-arrival times can be decomposed into simpler auxiliary variables that can be sampled directly, giving exact simulation with no approximation. We establish that the auxiliary variables provides information on the parent process for each event time. The algorithm correctness is shown by verifying the simulated intensities with their theoretical moments. A modular inference procedure consisting of Gibbs samplers through the auxiliary variable augmentation and adaptive rejection sampling is presented. Finally, we compare our proposed simulation method against existing methods, and find significant improvement in terms of algorithm speed. Our inference algorithm is used to discover the strengths of mutually excitations in real dark networks.

## 1. Introduction

Phenomena such as earthquakes, contagion (in diseases and economic senses) and consumers' buying behaviour, tend to occur in succession, and may be attributed to endogenous (internal) and/or exogenous (external) factors. For example, a disease epidemic may be concentrated in a geographic region, and for a period of time, the spread is endogenous since the occurrence of an infection increases the contagion rate for future infections. These examples exhibit clustering characteristics.

The advent of self-exciting processes to model clustering dates back to Hawkes (1971), who addressed the problem of assessing earthquake occurrences. What was new in his paper in contrast to other point processes was a concrete and mathematically tractable point process model with the inclusions of self-exciting and self-similarity behaviours (Hawkes and Oakes, 1974), meaning an occurrence of one event triggers a series of similar events.

This paper focusses on the simulation and inference of multidimensional Hawkes processes with exponential kernels. We follow the set-up of Brémaud and Massoulié (2002) and Dassios and Zhao (2011) to allow the level of excitations to be random. The random excitations can be seen as *marks*, which signify the level at which the intensity increases after the occurrence of an event (Daley and Vere-Jones, 2003, Section 6.4). These Hawkes processes are also called *marked* Hawkes processes. Additionally, we let the decay parameter of the exponential kernels to be flexible rather than fixed, like those in Muni Toke and Pomponio (2012). This allows a variety of curvature in the intensity process to be modelled. In this paper, we will henceforth refer the multidimensional Hawkes processes with random excitations and differing exponential decays simply as Hawkes processes.

The motivations for targeting the exponential kernel are twofold. Firstly, this kernel has seen a wide range of applicability due to its simplicity and intuitiveness. Secondly, the underlying intensities of the Hawkes processes at any particular time can be rewritten recursively Ozaki (1979). Because of this, we can achieve computational efficiency in both simulations and inferences.

### 1.1. Simulation Methods

Simulation of Hawkes processes generally falls into three categories. The first approach, proposed by Ozaki (1979), opts to sample the inter-arrival times of the events in Hawkes processes directly, by inverse sampling from their conditional cumulative probability density (CDF). Simulating the inter-arrival times can be computationally expensive, because solving the equation involving the inverse of the CDF requires numerical methods such as Newton's method and Brent's method.

The second way of simulating Hawkes processes is by *thinning* the samples from homogeneous Poisson processes and is akin to that of a rejection sampler. This method is known as Ogata's modified thinning algorithm (Ogata, 1981) as it is based on the thinning algorithm launched by Lewis and Shedler (1979) for the inhomogeneous Poisson processes. The thinning algorithm remains popular due to its flexibility and simplicity. Recently, Zaatour (2014) provides a fast implementation for simulating multivariate Hawkes processes based on Ogata's modified thinning algorithm; while Farajtabar et al. (2015) propose a variant of the thinning algorithm to simulate Hawkes processes on information diffusions of social networks. As with other rejection samplers, a drawback of thinning is that some samples are discarded.

The third simulation approach is known as the cluster based method as it uses the Poisson cluster process representation of Hawkes processes. Observing that each event time generates an inhomogeneous Poisson process, Brix and Kendall (2002) proposed a simulation algorithm that firstly samples event times (called immigrants) from a homogeneous Poisson process, then recursively samples more event times (known as offsprings) from the inhomogeneous Poisson processes associated with the existing event times. This method is extended by Møller and Rasmussen (2005), relaxing some requirements of Brix and Kendall (2002). An approximate but faster variant is given by Møller and Rasmussen (2006). An advantage of the cluster based approach over the other two is that the simulation also pro-vides the *triggering source* of the event times (also known as branching structure), useful in several applications including finance and social networks (Crane and Sornette, 2008; Farajtabar et al., 2015). The cluster based simulation algorithm is simple to implement, however, a naïve implementation requires the Hawkes processes to be stable (or stationary).

A notable simulation algorithm that does not belong to these categories is of Dassios and Zhao (2013). They adopt Ozaki's approach but sample the inter-arrival times directly without resorting to numerical methods; thus achieving fast simulation with no wastage.

### 1.2. Major Contributions

**Simulation.** This paper presents a new simulation method addressing some limitations of existing samplers. Our method generalises Dassios and Zhao (2013) through the superposition properties of point processes (Daley and Vere-Jones, 2003, Theorem 2.4.VI) rather than relying on the Markov property of the intensities formulated through an ordinary differential equations (ODE). With this, we are able to simulate multidimensional Hawkes processes that have *dissimilar* exponential decay kernels. Our method is similar to Ozaki (1979) in that we simulate the inter-arrival times directly, giving efficient sampler without *wastage*. Unlike Ozaki (1979), we do not resort to using any approximation techniques (hence *exact* simulation), instead, our algorithm recognises that the inter-arrival times can be formulated as first order statistics (David and Nagaraja, 2003) of simpler *auxiliary* variables that can be sampled easily and directly. Interestingly, although developed with different theory, our algorithm bares some resemblance to that of Dassios and Zhao (2013). One major different, however, is that our algorithm conveys information regarding the source that triggers the event times, which is currently only possible with cluster based methods.

**Stationarity conditions.** We generalise slightly the stationarity results of Bacry et al. (2015) and Hawkes (1971) by extending the unmarked multivariate Hawkes to handle random excitations $Y \in L^2$ (Hilbert space, $\mathbb{E}[Y^2] < \infty$). Precisely, we derive the stationary average intensities for multivariate Hawkes processes with *random* excitations and dissimilar decays. These are used to verify the correctness of our simulation algorithm. We note that although the stationary average intensities can be spelled out in closed form, their number of terms grows factorially with the dimensionality of the Hawkes processes.

**Inference.** A modular inference procedure consisting fully of Gibbs samplers is presented. We exploit the inherent branching structures for multidimensional Hawkes processes and augment the parameter space so that we can apply Gibbs sampling through adaptive rejection sampling (ARS, Gilks and Wild, 1992). The conditions for log-concavity tied to the posterior are spelled out. Here we demonstrate how such a strategy can be applied to multidimensional Hawkes in order to greatly improve inference efficiency.

We start Section 2 by introducing the Hawkes processes mathematically. Section 3 then extends the stability of the Hawkes processes to random excitations, and proceeds with the derivation of the stationary average intensities. In Section 4, we present our new simulation method for multivariate Hawkes processes. Section 5 discusses the MCMC algorithms that provide significant flexibility to do parameter estimation for Hawkes with dissimilar decays. Section 6 presents measures of simulation efficiency against existing samplers and assesses the proposed inference algorithm. We evaluate this Hawkes model empirically on a real world dataset of Dark-networks in Section 7. In Section 8, discussion on related inference methods ensues. Section 9 concludes.

## 2. Marked Multidimensional Hawkes Process

A Hawkes process may be *completely* characterised by its underlying intensity functions, or it may be formulated as a Poisson cluster process as illustrated by Hawkes and Oakes (1974) and Møller and Rasmussen (2005, 2006). Here, we focus on the intensity formulation (for details, see Daley and Vere-Jones, 2003).

Consider the $M$-dimensional Hawkes processes with dissimilar exponential decaying intensities, the respective intensity function (we assume time $t$ starts at 0) for each process $m \in \{1, \ldots, M\}$ can be written as

$$\lambda_m(t) = \mu_m + \sum_{i=1}^{M} \lambda_m^i(t), \qquad t > 0;\ m, i \in \{1, \ldots, M\}, \tag{1}$$

where $\mu_m > 0$ is the *background intensity* for process $m$, that is, the constant rate for which the events in process $m$ are generated. The $\lambda_m^i(t)$ inside the summation is the *additional* intensity corresponding to either self-excitation (when $i = m$) or external-excitation (when $i \neq m$), this added intensity is attributed to the events associated to process $i$. We assume the following general form for $\lambda_m^i(t)$:

$$\lambda_m^i(t) = \xi_m^i(t) + \int_0^t \kappa_m^i\big(t - s, Y_m^i(s)\big) \,\mathrm{d}N^i(s), \tag{2}$$

which is equivalent to the more commonly used summation form, given the event times $t_j^i$ for each process $i$:

$$\lambda_m^i(t) = \xi_m^i(t) + \sum_{j=1}^{N^i(t)} \kappa_m^i\big(t - t_j^i, Y_{m,j}^i\big)\, I(t \geq t_j^i). \tag{3}$$

Here, $I(\cdot)$ denotes the indicator function and $N^i(t)$ represents the $i$-th counting process of the multidimensional Hawkes, that is, $N^i(t)$ is the number of events attributed to process $i$ observed at and before time $t$. Also defined $Y_{m,j}^i$ is a short hand for $Y_m^i(t_j^i)$. On the other hand, $\kappa_m^i(t, y)$ is the kernel function of the Hawkes processes, while $\xi_m^i(t)$ corresponds to the intensity associated with the *edge effect* (see Møller and Rasmussen, 2005) generated by previously unobserved event times. They are assumed to follow an exponential decay:

$$\xi_m^i(t) = Y_m^i(0)\, e^{-\delta_m^i \times t}, \qquad \kappa_m^i(t, Y) = Y\, e^{-\delta_m^i \times t}, \tag{4}$$

where $Y$ denotes the non-negative jump size in the intensity function, which is assumed to be known or can easily be sampled, say, $Y \sim \mathrm{Gamma}(\alpha, \beta)$. A random $Y$ is also known as a mark. These Hawkes processes reduce to the ordinary Hawkes processes with exponential decay when the $Y$ is constant, for instance, see the model of Muni Toke and Pomponio (2012). We note that the decay rates $\delta_m^i$ are not assumed to be the *same* unlike in most models, for example, the decay rates in Dassios and Zhao (2013, Section 5) are constant within each process. It is also interesting to note that the multivariate Hawkes processes can be formulated as a shot noise Cox process by removing the self-excitation bit, easily achieved by setting certain $Y$ to be zero. For a more detailed reviews on the Hawkes processes, we refer the readers to the recent review papers by Bacry et al. (2015).

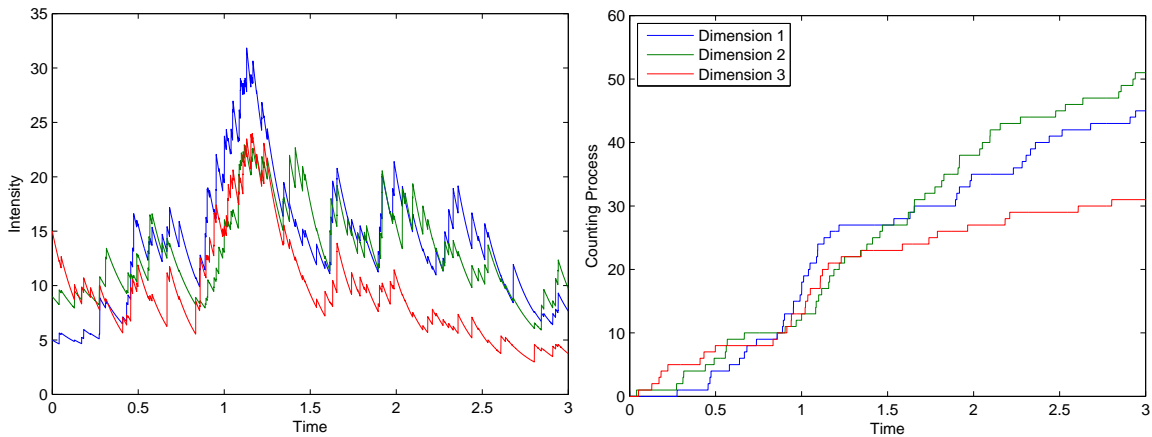

Figure 1: A sample of a three-dimensional Hawkes processes. The left plot graphs the realised intensity function $\lambda_m(t)$ for the Hawkes processes, while the right plot shows the corresponding counting processes $N^m(t)$. Each increase in the counting processes corresponds to a jump in each of the intensity functions.

As illustration, we present a realisation (or sample) of a three-dimensional Hawkes process in Figure 1. The Hawkes processes exhibit positive correlation due to mutual-excitation, which is observable from the plot of the intensity functions.

## 3. Stability of Hawkes Processes

A Hawkes process is *stable* or *stationary* if its expected intensity at any time is bounded (i.e., does not blow up), this also means that a *unique* stationary average intensity exists (Brémaud and Massoulié, 1996). For our Hawkes model described in Section 2, a sufficient condition for stationarity is that the *spectral radius* of the matrix $\boldsymbol{\Omega}$, with entries

$$\omega_{mi} = \int_0^\infty \mathbb{E}\Big[\,\big|\kappa_m^i\big(t, Y_m^i(\cdot)\big)\big|\,\Big]\,\mathrm{d}t\,, \qquad (5)$$

is strictly less than 1 (see Brémaud and Massoulié, 1996, Theorem 7; and Brémaud et al., 2002). The spectral radius of a matrix $\boldsymbol{\Omega}$ is defined as the maximum of the absolute value of the eigenvalues of $\boldsymbol{\Omega}$: denoting $\phi = \{\ldots, \phi_i, \ldots\}$ as the eigenvalues of $\boldsymbol{\Omega}$ (i.e., $\boldsymbol{\Omega}x = \phi_i x$ for some eigenvector $x$), the spectral radius is given as $\rho(\boldsymbol{\Omega}) = \max_i |\phi_i|$. Under the exponential kernels, the entry $\omega_{mi}$ can be simplified to $\omega_{mi} = \gamma_m^i/\delta_m^i$, where $\gamma_m^i = \mathbb{E}[Y_m^i(\cdot)]$ is the expected intensity jump. This quantity is a known constant since we have assumed $Y_m^i(\cdot)$ to be independent and identically distributed given $m$ and $i$.

In the following, we derive the stationary average intensities of multidimensional Hawkes. We note that the stationary average intensity for a univariate marked Hawkes process has been derived in the previous work of Brémaud and Massoulié (2002) and Dassios and Zhao (2013), while those of multivariate *unmarked* Hawkes processes were stated by Hawkes (1971). However, to the best of our knowledge, the stationary average intensities for multi-variate Hawkes processes with random excitations and dissimilar decays were never derived nor published, thus we fill in the gap.

### 3.1. Stationary Average Intensity for Hawkes Processes

If a Hawkes process is stationary, then the expectation of its intensity converges to its stationary average intensity $b_m$, that is, $\mathbb{E}[\lambda_m(t)] \to b_m$ when $t \to \infty$. Here, we derive the stationary average intensities $\mathbf{B} = [b_1, \ldots, b_M]^\mathsf{T}$ for $M$-dimensional Hawkes processes, provided their stationary condition is satisfied.

From the intensity function laid out in Equation (1), by expanding with Equation (2):

$$\lambda_m(t) = \mu_m + \sum_{i=1}^{M} \left[ Y_m^i(0) \, e^{-\delta_m^i t} + \int_0^t \kappa_m^i\bigl(t - s, Y_m^i(s)\bigr) \, \mathrm{d}N^i(s) \right], \tag{6}$$

if we take expectation on both side and then let $t \to \infty$, we have

$$\lim_{t \to \infty} \mathbb{E}[\lambda_m(t)] = \mu_m + \sum_{i=1}^{M} \lim_{t \to \infty} \mathbb{E}\left[ \int_0^t \kappa_m^i\bigl(t - s, Y_m^i(s)\bigr) \, \mathrm{d}N^i(s) \right]. \tag{7}$$

To evaluate the inner expectation in Equation (7), we first derive the expected value of $N^m(t)$, and then take derivative to obtain $\mathbb{E}[\mathrm{d}N^m(t)]$. Using law of iterated expectation, the expected value for $N^m(t) - N^m(0)$, where $N^m(0)$ is the number of events at time $t = 0$, commonly assumed to be $N^m(0) = 0$, is derived as

$$\mathbb{E}\bigl[N^m(t) - N^m(0)\bigr] = \mathbb{E}\Bigl[\mathbb{E}\bigl[N^m(t) - N^m(0) \,\big|\, \Lambda_m(t)\bigr]\Bigr] = \mathbb{E}\bigl[\Lambda_m(t)\bigr], \tag{8}$$

where $\Lambda_m(t) = \int_0^t \lambda_m(s) \, \mathrm{d}s$ is the compensator for process $m$, which can be computed as

$$\Lambda_m(t) = \mu_m t + \sum_{i=1}^{M} \left[ \frac{Y_m^i(0)}{\delta_m^i} \left(1 - e^{-\delta_m^i t}\right) + \sum_{j=1}^{N^i(t)} \frac{Y_{m,j}^i}{\delta_m^i} \left(1 - e^{-\delta_m^i (t - t_j^i)}\right) \right]. \tag{9}$$

Note that evaluation of the inner expectation in Equation (8) follows from the fact that $N^m(t) - N^m(0) \,|\, \Lambda_m(t)$ is Poisson distributed (at fixed $t$) with rate $\Lambda_m(t)$. Now, taking derivative on both side of Equation (8) with respect to $t$, we get

$$\mathbb{E}\bigl[\mathrm{d}N^m(t)\bigr] = \mathbb{E}\bigl[\lambda_m(t)\bigr] \mathrm{d}t. \tag{10}$$

Back to Equation (7), bringing the expectation into the integral, substituting $\mathbb{E}[\mathrm{d}N^i(s)]$ via Equation (10), and then replacing $\mathbb{E}[\lambda_i(s)]$ by its stationary average intensity $b_i$ gives

$$b_m = \mu_m + \sum_{i=1}^{M} \gamma_m^i \, b_i \lim_{t \to \infty} \int_s^t e^{-\delta_m^i (t - s)} \, \mathrm{d}s = \mu_m + \sum_{i=1}^{M} \omega_{mi} \, b_i. \tag{11}$$

Rewriting the system of equations of above (Equation 11) in matrix form, and solving for $\mathbf{B} = [b_1, \ldots, b_M]^\mathsf{T}$ leads to $\mathbf{B} = (\mathbf{I} - \mathbf{\Omega})^{-1}\mu$, where $\mathbf{I}$ is an $M \times M$ identity matrix, $\mathbf{\Omega}$ is a matrix of $\omega_{mi}$ as defined in Equation (5), and $\mu = (\mu_1, \ldots, \mu_M)^\mathsf{T}$ is a vector of background intensities. We note that $\mathbf{B}$ exists whenever the matrix $\mathbf{I} - \mathbf{\Omega}$ is invertible. In Proposition 1, we show that this matrix is invertible when the stationary condition is satisfied. Then, we summarise our result of this section in Proposition 2 and state some corollaries.

**Proposition 1** *If the spectral radius of a matrix $\boldsymbol{\Omega}$ is strictly less than 1, then the matrix $\mathbf{I} - \boldsymbol{\Omega}$ is invertible.*

*Proof:* First note that the eigenvalues of the matrix $\boldsymbol{\Omega}$, denoted by $\{\phi_i\}$, satisfy the equation $\det(\boldsymbol{\Omega} - \phi_i \mathbf{I}) = 0$. If the spectral radius of $\boldsymbol{\Omega}$ is strictly less than 1, that is, $\max_i\{|\phi_i|\} < 1$, then all $\phi_i$ satisfy $|\phi_i| < 1$.

Now, notice that for a constant $k = 1$, $\det(\boldsymbol{\Omega} - k\mathbf{I}) \neq 0$. This can be proven by contradiction: If $\det(\boldsymbol{\Omega} - k\mathbf{I}) = 0$ for $k = 1$, then $k$ is an eigenvalue and $k \in \{\phi_i\}$. However, since all $|\phi_i| < 1$, $k$ cannot be 1, which is a contradiction. It follows that $\det(\mathbf{I} - \boldsymbol{\Omega}) = -\det(\boldsymbol{\Omega} - k\mathbf{I}) \neq 0$, implying that $\mathbf{I} - \boldsymbol{\Omega}$ is invertible. ∎

**Proposition 2 (Stationary average intensity)** *If an $M$-dimensional marked Hawkes process with random excitations and differing exponential decay kernels described in Section 2 is stationary, then its stationary average intensity is given by $\mathbf{B} = (\mathbf{I} - \boldsymbol{\Omega})^{-1}\boldsymbol{\mu}$. The existence of $\mathbf{B}$ is guaranteed by Proposition 1.*

**Corollary 3** *In the special case of $M = 1$, the stationary average intensity $\mathbf{B} = b_1$ is*

$$b_1 = \mu_1/(1 - \omega_{11}). \tag{12}$$

This is consistent with Brémaud and Massoulié (2002) and Dassios and Zhao (2013).

**Corollary 4** *For the special case of $M = 2$, the stationary average intensity $\mathbf{B} = [b_1, b_2]^\mathsf{T}$ can be penned down as*

$$b_1 = \frac{\mu_1(1 - \omega_{22}) + \mu_2\, \omega_{12}}{(1 - \omega_{11})(1 - \omega_{22}) - \omega_{12}\, \omega_{21}}, \qquad b_2 = \frac{\mu_2(1 - \omega_{11}) + \mu_1\, \omega_{21}}{(1 - \omega_{11})(1 - \omega_{22}) - \omega_{12}\, \omega_{21}}. \tag{13}$$

We note that the explicit expression for $\mathbf{B}$ becomes intractable to write down as the dimension $M$ increases, since the number of terms grows factorially.

## 4. Exact Simulation of Multidimensional Hawkes

Here, we detail our exact (no approximation) simulation procedure that exploits the superposition property (Daley and Vere-Jones, 2003) of point processes. Our discussion will focus on the general $M$-dimensional Hawkes, though, we note that the simulation process for the special case of univariate Hawkes is presented in Appendix A in the supplementary.

Define $\mathbf{r} = \{\ldots, r_j, \ldots\}$ as the sorted times for all events in the Hawkes processes. Also, let $N(t) = \sum_m N^m(t)$ be the total number of events occur until time $t$. Mathematically, $\mathbf{r}$ relates to $\mathbf{t}^m = \{\ldots, t_k^m, \ldots\}$, $m = 1, \ldots, M$, via

$$\mathbf{r} = \text{sort } \cup_m \cup_k t_k^m \quad \text{(ascending)}. \tag{14}$$

Due to the coupling from mutual excitation, it is difficult to sample the event time $t_k^m$ in each process $m$ directly. Thus, we instead sample the event times $r_j$ sequentially conditioned on $r_1, \ldots, r_{j-1}$, and then recover the event times $\mathbf{t}^m$ from $\mathbf{r}$. Employing the superposition theory, the intensity for counting process $N(t)$ can be written as $\lambda(t) = \sum_m \lambda_m(t)$. For this process, the CDF for the inter-arrival time, $d_j = r_j - r_{j-1}$, is thus (see Ozaki, 1979)

$$F_{d_j}(s) = P(d_j < s) = 1 - \exp\left(-\int_{r_{j-1}}^{r_{j-1}+s} \lambda(t)\,\mathrm{d}t\right), \qquad s > 0, \tag{15}$$

which is difficult to sample directly. However, by noticing that

$$\left(1 - F_{d_j}(s)\right) = \prod_{m=1}^{M} \prod_{i=0}^{M} \left(1 - F_{a^i_{mj}}(s)\right), \tag{16}$$

where $F_{a^i_{mj}}(s)$ is the CDF of the inter-arrival time $a^i_{mj}$ generated by the intensity $\lambda^i_m(t)$ (define $\lambda^0_m(t) = \mu_m$ for convenience), we can recast $d_j$ as a first order statistic (David and Nagaraja, 2003), as follows:

$$d_j = \min_{m,i} a^i_{mj}. \tag{17}$$

Hence, we can obtain $d_j$ by first sampling all $a^i_{mj}$ and then pick the minimum. Sampling $a^0_{mj}$ is trivial since $a^0_{mj} \sim \text{Exp}(\mu_m)$. While simulating $a^i_{mj}$ for $i \neq 0$ can be done using the inverse CDF method, since its CDF

$$F_{a^i_{mj}}(s) = 1 - \exp\left(-\int_{r_{j-1}}^{r_{j-1}+s} \lambda^i_m(t)\,dt\right) = 1 - \exp\left(-\frac{1}{\delta^i_m} \lambda^i_m(r_{j-1}) \cdot \left(1 - e^{-\delta^i_m s}\right)\right) \tag{18}$$

can now be inverted. The expression in the exponent of Equation (18) is obtained using the recursive formula from Ozaki (1979), and that $\lambda^i_m(r_{j-1})$ is known and cached. We note that this CDF corresponds to a defective random variable (Feller, 1971, Chapter VI), which has a *probability mass* at $\infty$. This corresponds to the case where an event *never* arrive:

$$P(a^i_{mj} = \infty) = 1 - \lim_{s \to \infty} F_{a^i_{mj}}(s) = \exp\left(-\frac{1}{\delta^i_m} \lambda^i_m(r_{j-1})\right) \tag{19}$$

To sample $a^i_{mj}$, we generate $u \sim \text{U}(0,1)$, if $u < 1 - \exp\left(-\frac{1}{\delta^i_m} \lambda^i_m(r_{j-1})\right)$, we let

$$a^i_{mj} = -\frac{1}{\delta^i_m} \log\left(1 - \frac{\delta^i_m}{\lambda^i_m(r_{j-1})} \log(1-u)\right), \tag{20}$$

otherwise, we set $a^i_{mj} = \infty$.

After obtaining $d_j$ with Equation (17), we update the event time $r_j = r_{j-1} + d_j$. We note that the smallest of $a^i_{mj}$ provides us additional information, which is captured by the auxiliary variables $(Z_j, X_j) = (m^*, i^*) = \arg\min_{m,i} a^i_{mj}$, where $Z_j$ is the type of $r_j$ (i.e., which process it is in), and $X_j$ tells us the origin of $r_j$, that is, from which process it comes from. To illustrate, if $X_j = 0$ then $r_j$ is an immigrant, if $Z_j = X_j$ then $r_j$ arises from self-excitation, otherwise $r_j$ is caused by an external process $i^*$.

We proceed by sampling $Y^i_m(r_j)$ for $i = m^*$ and $m = 1, \ldots, M$, which are the intensity jumps added to each process generated by $r_j$. The update rule for the intensity cache ($m = 1, \ldots, M; i = 1, \ldots, M$) is given as follows:

$$\lambda^i_m(r_j) = \lambda^i_m(r_{j-1}) e^{-\delta^i_m d_j} + Y^{m^*}_{m,j} I(i = m^*). \tag{21}$$

We then increase the counting process $N^{m^*}(r_j)$ by one, note that this also means $t^{m^*}_k = r_j$ for $k = N^{m^*}(r_j)$. This process is repeated until the maturity time $T$, that is, until $r_j > T$. Since this last event $r_j$ is outside the observation window, we discard it and keep the others as samples. We summarise the full simulation procedure in Algorithm 1.

For completeness, we discuss the differences between our work and that of Dassios and Zhao (2013) through the simulation of a bivariate Hawkes. This is presented in Appendix B.

**Algorithm 1** Exact Simulation of Multidimensional Hawkes

Given Input $M$, $\mu_m$, $Y_m^i(0)$, $\delta_m^i$, $N^m(0)$, $T$, and distribution of $Y_{m,j}^i$.

1. Initialise $r_0 = 0$.
2. For $m = 1, \ldots, M$, $i = 1, \ldots, M$: set $\lambda_m^i(r_0) = Y_m^i(0)$.
3. For $j = 1, 2, 3, \ldots$:
   (a) For $m = 1, \ldots, M$:
       i. Sample $a_{mj}^0 \sim \text{Exp}(\mu_m)$.
       ii. For $i = 1, \ldots, M$: sample $a_{mj}^i$ using inverse CDF method (Equation 18).
   (b) Set $r_j = r_{j-1} + \min_{m,i} a_{mj}^i$.
   (c) Set $(Z_j, X_j) = (m^*, i^*) = \arg\min_{m,i} a_{mj}^i$.
   (d) For $m = 1, \ldots, M$:
       i. Sample $Y_{m,j}^{m^*}$.
       ii. For $i = 1, \ldots, M$: update $\lambda_m^i(r_j)$ according to Equation (21).
       iii. Update $N^m(r_j) = N^m(r_{j-1}) + I(m = m^*)$.
   (e) Set $t_k^{m^*} = r_j$ for $k = N^{m^*}(r_j)$.
   (f) Terminate loop if $r_j > T$, discard $r_j$ and its associated variables.

## 5. Parameter Inference from Observed Data

In practice, the values of the parameters are not known and thus will need to be inferred from the data. Here, we briefly discuss the maximum likelihood estimation (MLE) method, which is commonly used for parameter estimation. For example, Ozaki (1979) and Bowsher (2007) use the MLE to learn the parameters of Hawkes processes with constant level of excitations. However, in Section 6, we find that the MLE tends to exhibit positive bias for the Hawkes model. Thus, we propose a Markov chain Monte Carlo (MCMC) algorithm for performing fully Bayesian inference on the parameters. Our inference method follows the principle of Rasmussen (2013) but consists fully of Gibbs samplers by employing adaptive rejection sampling (ARS, Gilks and Wild, 1992) and auxiliary data augmentation (van Dyk and Meng, 2001). Conditions for which the ARS can be applied are outlined.

### 5.1. Maximum Likelihood Estimation on Parameters

The MLE has frequently been employed in practice for parameter estimation of statistical models. The MLE is simple to apply since one would only need to know the likelihood function associated with the statistical model.

The maximum likelihood estimates are obtained by selecting the parameter values that maximises the model's likelihood function or, equivalently, its log. The joint log likelihood for the Hawkes processes described in Section 2 can be derived as

$$\log P(\mathbf{t}, \mathbf{Y} \mid \cdots) = \sum_{m=1}^{M} \left[ \sum_{j=1}^{N^m(T)} \left( \log \lambda_m(t_j^m) \right) - \Lambda_m(T) \right] + \log P(\mathbf{Y} \mid \cdots) \qquad (22)$$

where $\mathbf{t}$ and $\mathbf{Y}$ represent a collection of all $t_j^m$ and $Y_{m,j}^i$, while $\Lambda_m(t)$ is the compensator defined in Equation (9), and $\log P(\mathbf{Y} \mid \cdots)$ is the log likelihood of the distribution $\mathbf{Y}$.

### 5.2. Markov Chain Monte Carlo Algorithm for Bayesian Inference

Bayesian inference analyses the marginal posterior distributions of the parameters of interest. Standard approach involves deriving the posterior distributions using the Bayes rule and then calculating the statistics of the posterior distributions. However, performing exact Bayesian inference on nonparametric models is often intractable due to the difficulty in deriving the closed-form posterior distributions. This motivates the use of MCMC methods for approximate inference. Refer to Gelman et al. (2013) for a detailed discussion on these.

Due to space, technicality of our proposed MCMC algorithm will be detailed in Appendix C. Here, we outline the basics. For simplicity, we assume the levels of excitation $Y_{m,j}^i$ follows i.i.d. gamma distribution given by

$$Y_{m,j}^i \sim \text{i.i.d. Gamma}\left(\alpha_m^i, \beta_m^i\right), \qquad \text{for } j = 1, \ldots, N^i(T), \qquad (23)$$

though noting that the distributions can easily be modified or extended, say, following a stochastic differential equation (SDE, see Lee et al., 2016). One benefit of choosing the gamma distribution is that the marginal posterior of $Y_{m,j}^i$ will then be conjugate gamma.

We denote $\boldsymbol{\Theta} = \{\mu_m, \delta_m^i, \alpha_m^i, \beta_m^i\}$ as the set of all parameters of interest. These parameters are assigned gamma prior either due to conjugacy (such that their posterior is also gamma), or simply for convenience. We note that this choice also allows the non-conjugate variables to exhibit log-concavity for ARS. Thus, our MCMC algorithms will be fully Gibbs samplers. The joint posterior for all the parameters can be derived as

$$P(\boldsymbol{\Theta} \mid \mathbf{t}, \mathbf{Y}) \propto P(\boldsymbol{\Theta}) \prod_{m=1}^{M} \left( \prod_{j=1}^{N^m(T)} \lambda_m(t_j^m) \right) e^{-\Lambda_m(T)} \prod_{i=1}^{M} \prod_{j=1}^{N^i(T)} (Y_{m,j}^i)^{\alpha_m^i - 1} e^{-\beta_m^i Y_{m,j}^i} \qquad (24)$$

where $P(\boldsymbol{\Theta})$ is the prior likelihood. The relevant marginalised posteriors can then be obtained by dropping the irrelevant terms. Once we have the posteriors, we can proceed with Gibbs sampling. Appendix C outlines the details.

**Log-concavity of posteriors.** Adaptive rejection sampling (ARS) is a method for efficiently sampling from densities which are log-concave. It is extremely useful in applications of Gibbs sampling, where full-conditional distributions are algebraically involved yet often log-concave (Gilks and Wild, 1992). To that end, we use the ARS to infer parameters for our multidimensional Hawkes process model, in particular $\delta$ and $\alpha$. We shall detail our workings for $\delta$ and similar lines can be replicated for $\alpha$ (see Appendix C). The posterior for

$\delta$ can be computed as $P(\delta_m^i \mid \mathbf{t}, \mathbf{Y}, \mathbf{A}, \tau, \psi) \propto$

$$\left(\delta_m^i\right)^{\tau_{\delta_m^i}-1} \exp\left[-\delta_m^i \left(\psi_{\delta_m^i} + \sum_{j=1}^{N^m(T)} \sum_{k=0:\, t_k^i < t_j^m}^{N^i(T)} (A_j^m)_{ik}(t_j^m - t_k^i)\right) - \sum_{k=0}^{N^i(T)} \frac{Y_{m,k}^i}{\delta_m^i}\left(1 - e^{-\delta_m^i(T - t_k^i)}\right)\right] \quad (25)$$

where we have introduced $Y_{m,0}^i$ as a short hand for $Y_m^i(0)$ and $t_0 = 0$ to simplify the posterior. We prove that whenever the shape parameter $\tau_{\delta_m^i} > 1$, the posterior is log-concave and thus the ARS method can be used. To see this, first note that the function

$$f(\delta_m^i) = \left[2 - \left(\left(\delta_m^i(T - t_k^i) + 1\right)^2 + 1\right) e^{-\delta_m^i(T - t_k^i)}\right] \quad (26)$$

is strictly positive for all $\delta_m^i > 0$. The proof is as follows. We show that the left limit is

$$\lim_{\delta_m^i \to 0^+} f(\delta_m^i) = 2 - \left((0 + 1)^2 + 1\right) e^0 = 2 - 2 = 0 \quad (27)$$

and that the function $f(\delta_m^i)$ is monotonically increasing by showing that its gradient (derivative) is positive for all $\delta_m^i > 0$:

$$f'(\delta_m^i) = (\delta_m^i)^2 (T - t_k^i)^3 \, e^{-\delta_m^i(T - t_k^i)} > 0 \quad (28)$$

since $(T - t_k^i) > 0$. We have shown that $f(\delta_m^i)$ is strictly positive for all $\delta_m^i > 0$. ∎

## 6. Assessments and Comparisons

We perform some experiments to assess the correctness of our proposed sampler. In particular, we (1) present some statistics on the simulated synthetic data; (2) verify our algorithm against the theoretical stationary average intensities given in Proposition 2; (3) compare the proposed sampler against several existing methods; (4) perform MLE and MCMC methods to recalibrate the ground truth parameters used in simulations; and (5) motivate the Hawkes model with an application. We note that our proposed simulation algorithm is implemented in MATLAB to achieve speed up by exploiting vectorised computation instead of using loops. The source code for the simulation algorithm is made available online.

### 6.1. Simulation Statistics

In the following experiments, we curated a set of ground truth parameters for simulations and verification of our proposed sampler. The parameters are chosen for illustration, and we note that the same conclusion can be drawn from using another sets of parameters, albeit not explicitly shown here. We focus on three-dimensional Hawkes processes, this strikes a balance between complexity and ease of presentation. For simplicity, we fix the value of $Y_m^i(0)$ to $[(1,0,1)^\mathsf{T}; (1,1,5)^\mathsf{T}; (5,1,8)^\mathsf{T}]^\mathsf{T}$, and assume that the intensity jump size $Y_{m,j}^i$ follows a gamma distribution with $\alpha_m^i$ and $\beta_m^i$ as the shape and rate parameters.

A simulated path with this set of parameters is presented in Figure 1, where we restrict the x-axis to $t = [0, 3]$ to emphasise the connection between the counting processes and their intensity. They are positively correlated, showing sign of mutually-excitation.

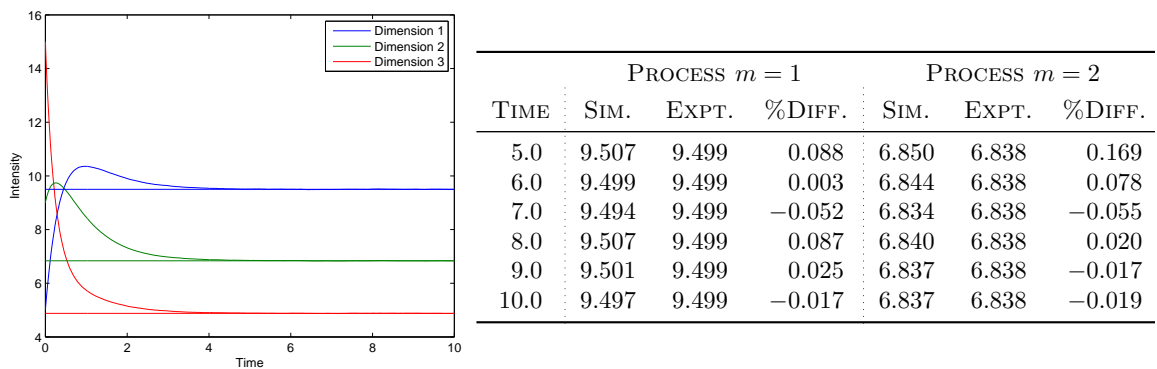

|  | Process $m=1$ | | | Process $m=2$ | | |
|---|---|---|---|---|---|---|
| Time | Sim. | Expt. | %Diff. | Sim. | Expt. | %Diff. |
| 5.0 | 9.507 | 9.499 | 0.088 | 6.850 | 6.838 | 0.169 |
| 6.0 | 9.499 | 9.499 | 0.003 | 6.844 | 6.838 | 0.078 |
| 7.0 | 9.494 | 9.499 | $-0.052$ | 6.834 | 6.838 | $-0.055$ |
| 8.0 | 9.507 | 9.499 | 0.087 | 6.840 | 6.838 | 0.020 |
| 9.0 | 9.501 | 9.499 | 0.025 | 6.837 | 6.838 | $-0.017$ |
| 10.0 | 9.497 | 9.499 | $-0.017$ | 6.837 | 6.838 | $-0.019$ |

Figure 2: Plot of simulated mean intensities vs the theoretical stationary average intensities of the three-dimensional Hawkes processes. The simulated means are represented by the curvy lines while the stationary average intensities are shown by the horizontal lines. From this plot, we can see that the mean intensities converge to their theoretical stationary value. They are almost indistinguishable for $t \geq 5$, thus validating the correctness of the proposed simulation algorithm.

### 6.2. Assessment with Stationary Average Intensities

One way to assess the correctness of our proposed method is by comparing the simulations against theoretical quantities. Here, we compare the mean of the simulated intensities against the stationary average intensities derived in Section 3. The simulated means are obtained by averaging the intensity functions from one million samples of three-dimensional Hawkes processes. They are presented in Figure 2, together with the stationary average intensities. Additionally, their deviations at times $t \geq 5$ are displayed. We find that their differences do not exceed by 0.2 %, which verifies the correctness of our algorithm.

### 6.3. Comparison with Existing Samplers

While the above experiments verify the correctness of the proposed simulation method, it is also important to compare against existing samplers. For multidimensional Hawkes, we compare against the R package "Hawkes" (Zaatour, 2014), which is written in C++ and integrated into R with Rcpp thus performs quickly. For this comparison, we let dimension $M = 50$, we set $Y_m^i(0)$ to zero and fix the intensity jump size to a constant $(Y_m^i(s) = 0.8)$ to accommodate the R package. For simplicity and to ensure stationarity, we set background intensity $\mu_m = 0.5$, and decay rate $\delta_m^i = 50$, also noting that the decay rates in Zaatour (2014) are not as flexible as ours. The maturity time is $T = 100$. We present the comparison result, computed from 1,000 runs, in Table 1.

In addition, through the simulation of univariate Hawkes processes, we perform comparison against the inverse sampling method of Ozaki (1979) and the cluster based method of Brix and Kendall (2002). Since implementation on these methods were not readily available, we implement them based on the algorithms outlined in the respective articles, thus might not be fully optimised, however, the same can be said to the implementation of our own sampler. The simulation settings (subscripts $m$ and $i$ suppressed) are $\mu = 0.5$, $Y_0 = 0$, $\delta = 1$, and $Y_s$ is fixed to 0.8. Different from the multivariate comparison, we set $T = 100,000$

Table 1: Comparison of the proposed algorithm against existing samplers. We compare the average number of events generated per simulation and its standard deviation, the numbers agree thus validating our algorithm. We then compare the computational time taken per simulation [second], its standard deviation [second], and the time taken per event time [microsecond]. We demonstrate that our sampler perform much faster compared to the others.

| Sampler | Events | StDev | Time/Sim [s] | StDev [s] | Time/Event [μs] |
|---:|---:|---:|---:|---:|---:|
| **Multivariate ($M = 50$)** | | | | | |
| Zaatour (2014) | 12 503 | 566 | 4.24 | 0.29 | 339 |
| Our sampler | 12 492 | 530 | 2.47 | 0.14 | **198** |
| **Univariate ($M = 1$)** | | | | | |
| Ozaki (1979) | 249 995 | 2505 | 32.73 | 0.69 | 131 |
| Brix & Kendall (2002) | 249 926 | 2544 | 14.31 | 1.09 | 57 |
| Our sampler | 249 891 | 2533 | 3.12 | 0.26 | **13** |

*Our sampler is equivalent to Dassios and Zhao (2013) for univariate Hawkes, thus not compared.

to eliminate the interference from overhead cost. The comparison result is shown in the second half of Table 1. All experiments in this section are performed on a machine with *Intel Core i7-2600 CPU @ 3.4GHz* and *4GB RAM*.

### 6.4. Recalibrating the Ground Truth Parameters

Here, we compare our proposed MCMC algorithm against the MLE, evaluated on a synthetic two-dimensional Hawkes data. Testing on the synthetic data allows us to gauge how well the MCMC algorithm and the MLE perform.

The results are obtained from 500 samples of simulated Hawkes processes. The MLE is obtained using a bounded constrained optimiser built upon MATLAB function *fminsearch*, which employs the Nelder-Mead simplex method. The MCMC estimates are acquired by performing the Gibbs samplers outlined in Section 5.2. These results, presented in Table 2, suggests superior performance of the proposed MCMC method in estimating the ground truth parameters, while the MLE of $\delta_m^i$ shows considerable positive bias.

## 7. Modelling Forum Posts from DarkNets

As an example of how multidimensional Hawkes processes with dissimilar decays may discover interpretable latent strengths of mutually excitations in the real world applications, we study the forum posts from DarkNets collected by National ICT of Australia during the period April 2014 through to August 2015. We employ our MCMC algorithm to learn the parameters and to assess the mutually excitatory elements of the forum posts.

**DarkNets.** Dark Networks or DarkNets refer to various illegal and covert social networks (Kaza et al., 2007). These networks are specifically designed to conceal the identity and location of their users. DarkNets provide users with access to illicit items through crypto-based transactions and discussion between users. Typical items are pharmaceuticals, narcotics, and fraudulent identity materials. Networks comprise a "market" and a "forum". The forum offers users the opportunity to discuss trade mechanisms, feedback on sellers and quality of product, and anything else that comes to mind. The market offers open listings of products for sale as well as feedback on sellers.

Table 2: Comparison of learned parameters against their ground true values. The parameters are learned from 500 simulated Hawkes processes. Majority of the MLE and MCMC estimates are very close to their true values. However, the MLE of the decay rates $\delta_m^i$ appears to exhibit positive bias. Overall, the MCMC estimates achieve a lower mean square error (MSE) compared to the MLE.

| Name | Var. | Process $m=1$ | | | Process $m=2$ | | |
|---|---|---|---|---|---|---|---|
| | | True | MLE | MCMC | True | MLE | MCMC |
| Background intensity | $\mu_m$ | 2.0000 | 2.0078 | 1.9026 | 1.0000 | 1.0051 | 0.8555 |
| Decay rates | $\delta_m^1$ | 6.0000 | 6.5367 | 6.0978 | 3.0000 | 4.0671 | 3.0790 |
| | $\delta_m^2$ | 2.0000 | 2.6464 | 2.4649 | 5.0000 | 5.4443 | 5.2633 |
| Shape parameters | $\alpha_m^1$ | 4.0000 | 4.0171 | 4.0293 | 1.0000 | 1.0103 | 1.0076 |
| | $\alpha_m^2$ | 2.0000 | 2.0135 | 2.0100 | 6.0000 | 6.0907 | 6.0638 |
| Rate parameters | $\beta_m^1$ | 2.0000 | 1.9996 | 2.0193 | 4.0000 | 4.0262 | 4.0407 |
| | $\beta_m^2$ | 5.0000 | 4.9969 | 5.0426 | 3.0000 | 3.0223 | 3.0351 |
| Mean square error | MSE | 0.0000 | 0.1009 | **0.0340** | 0.0000 | 0.1922 | **0.0148** |

**Data.** We consider two DarkNets data sets. They are the times at which forum chats are posted. The communication in the forum chats are related to the discussions of buying and selling Methamphetamine and Cannabis, respectively. Specifically, we treat the time stamps as event times and we would like to model the occurrence of forum chats.

**Modelling strength of mutual excitations.** We show that Hawkes processes are ideal for modelling the occurrence of forum posts in DarkNets. The model postulates that the occurrence of a forum chat will influence the rate at which future forum chats will arrive. In addition, we demonstrate that Hawkes processes can infer the strength of cross excitations, i.e., if the frequency one chats in Methamphetamine influences Cannabis and vice versa.

Using the method described in Section 5.2, we calibrated a two-dimensional Hawkes model to fit the DarkNets data. In this application, we are interested in analysing the expected levels of self and mutual excitation $\mathbb{E}[Y_{m,\cdot}^i]$. We find that the background intensity for the cannabis' forum ($\hat{\mu}_1 = 0.6063$) is much higher as a result of the observation of higher posts on the cannabis forum, while the background intensity for the methamphetamine's forum is much lower ($\hat{\mu}_2 = 0.0930$). Additionally, $\alpha_m^i$ and $\beta_m^i$ describe the distribution of the levels of excitation. In this case, the expected levels of excitation are found to be

$$\hat{\mathbb{E}}[Y_{1,\cdot}^1] = 0.7878, \qquad \hat{\mathbb{E}}[Y_{2,\cdot}^1] = 0.0995, \qquad \hat{\mathbb{E}}[Y_{1,\cdot}^2] = 0.2007, \qquad \hat{\mathbb{E}}[Y_{2,\cdot}^2] = 0.0808.$$

From the levels of excitations, we can say that the cannabis' forum posts have relatively high self-excitation, as shown by $\hat{\mathbb{E}}[Y_{1,\cdot}^1]$. While the levels of self-excitation for the methamphetamine's forum is relatively much lower. On the other hand, the levels of mutual excitation on both side directions are not very high, though still a significant amount.

## 8. Related Inference Techniques

We have proposed an MCMC inference method as an alternative to the MLE. The MLE, albeit simple and fast, was shown to exhibit positive bias for Hawkes processes. Similar positive bias is also observed in the experiments of Ozaki (1979). The MCMC methods, on the other hand, produce samples that are as if drawn from the posteriors provided that the algorithm has converged and that the number of samples are large. Our method employs fully Gibbs samplers via the ARS and auxiliary variable augmentation to improve the convergence. The MCMC methods are also used in learning a univariate Hawkes (Rasmussen, 2013) and a network Hawkes process (Linderman and Adams, 2014).

Besides the MCMC methods, several existing work uses expectation-maximisation (EM) based algorithm for parameters inference on Hawkes (Veen and Schoenberg, 2008; Simma and Jordan, 2010; Lewis and Mohler, 2011). The EM algorithm iteratively improves the likelihood function of the parameters by alternating between an expectation (E) step and maximisation (M) step. Extension on the EM algorithm includes the stochastic EM (Iwata et al., 2013) and the variational Bayes EM (Cho et al., 2014) which aim to improve the estimates. Alternatively, Zhou et al. (2013) adopt a majorisation-minimisation algorithm.

Notably, Linderman and Adams (2015) propose a stochastic variational inference algorithm for Hawkes processes, shown to achieve a tenfold speed up compared to a Gibbs sampler and a variational Bayes method. Finally, we note that Da Fonseca and Zaatour (2014) utilise generalised method of moments for learning a univariate Hawkes. This method is extremely fast, though only applied to Hawkes processes that satisfy Markovian constraint.

## 9. Conclusion

In this paper, we presented a novel simulation algorithm for multidimensional Hawkes processes with random excitations and exponential kernels with differing decay rates. By employing an auxiliary variable sampling mechanism developed from the superposition theory of point processes, our algorithm gives *exact* (without resorting to approximation) and efficient (no wastage) simulation. Moreover, our simulation method informs us the triggering source of an event time, that is, *which* Hawkes process that triggers the event.

An inference procedure consisting of fully Gibbs samplers through adaptive rejection sampling is presented. We exploited the inherent branching structures for multidimensional Hawkes processes and augment the parameter space. The conditions for log-concavity tied to the posteriors is spelled out. Our algorithm performs well on a synthetic setting and on a real world application.

## Appendix A. Simulation of Univariate Hawkes Process

Here, we describe the special case of simulating a univariate ($M = 1$) Hawkes process. We will suppress the subscripts $m$ and $i$ for clarity, for instance,

$$\mu := \mu_1, \quad \delta := \delta_1^1, \quad Y_j := Y_{1,j}^1, \quad t_j := t_j^1, \quad \lambda^1(t) := \lambda_1^1(t), \quad N(t) := N^1(t). \tag{29}$$

A sample of the Hawkes process is obtained by simulating the the sequence of *inter-arrival* times $a_j = t_j - t_{j-1}$ from their conditional cumulative probability density (CDF) directly. We define $t_0 = 0$ as the starting time, note that $t_0$ is not an event time albeit sharing the same symbol.

From Ozaki (1979), the CDF of the inter-arrival time $a_j$, derived from its conditional hazard function, is given as

$$F_{a_j}(s) = 1 - \exp\left(-\int_{t_{j-1}}^{t_{j-1}+s} \lambda(t)\,dt\right), \qquad s > 0. \tag{30}$$

This expression is difficult to *inverse* for sampling purposes, which leads to the use of numerical approximations (see, for example, Ozaki, 1979). However, if we exploit the superposition property of point processes (Daley and Vere-Jones, 2003), we see that:

1. If $t_j$ is generated from the intensity $\mu$ (i.e., $t_j$ is an immigrant) then the CDF of its inter-arrival time $a_j^0$ follows

$$\begin{aligned} F_{a_j^0}(s) &= 1 - \exp\left(-\int_{t_{j-1}}^{t_{j-1}+s} \mu\,dt\right) \\ &= 1 - \exp(-\mu s)\,, \end{aligned} \tag{31}$$

which is the CDF of an exponential distribution with rate parameter $\mu$.

2. Otherwise, $t_j$ is generated from $\lambda^1(t)$ (i.e., $t_j$ is an offspring), the CDF of its inter-arrival time $a_j^1$ is thus

$$F_{a_j^1}(s) = 1 - \exp\left(-\int_{t_{j-1}}^{t_{j-1}+s} \lambda^1(t)\,dt\right). \tag{32}$$

The value $\lambda^1(t)$ is defined in Equation (3), and the integral in Equation (32) can be simplified using the recursive formula from Ozaki (1979), displayed below:

$$\begin{aligned} \int_{t_{j-1}}^{t_{j-1}+s} \lambda^1(t)\,dt &= \int_{t_{j-1}}^{t_{j-1}+s} \sum_{k=0}^{N(t_{j-1})} Y_k\, e^{-\delta(t-t_k)}\,dt \\ &= \sum_{k=0}^{N(t_{j-1})} -\frac{Y_k}{\delta}\left(e^{-\delta(t_{j-1}+s-t_k)} - e^{-\delta(t_{j-1}-t_k)}\right) \\ &= \frac{1}{\delta}\left[\sum_{k=0}^{N(t_{j-1})} Y_k\, e^{-\delta(t_{j-1}-t_k)}\right]\left(1 - e^{-\delta s}\right) \\ &= \frac{1}{\delta}\lambda^1(t_{j-1}) \times \left(1 - e^{-\delta s}\right). \end{aligned} \tag{33}$$

Both of the above CDFs for the auxiliary variables $a_j^0$ and $a_j^1$ can be inverted, allowing sampling with the inverse CDF method. Sampling $a_j^0$ is simple, which is achieved by first drawing a number $v$ from standard uniform distribution $U(0,1)$, then let

$$a_j^0 = -\frac{1}{\mu}\log(1-v). \tag{34}$$

On the other hand, the CDF of $a_j^1$ corresponds to a defective random variable (Feller, 1971, Chapter VI), which has a *probability mass* at $\infty$, with probability $\exp\left(-\frac{1}{\delta}\lambda^1(t_{j-1})\right)$. Hence special consideration is needed, to sample $a_j^1$, we generate $u \sim U(0,1)$, if $u < 1 - \exp\left(-\frac{1}{\delta}\lambda^1(t_{j-1})\right)$, we let

$$a_j^1 = -\frac{1}{\delta}\log\left(1 - \frac{\delta}{\lambda^1(t_{j-1})}\log(1-u)\right), \tag{35}$$

otherwise, we set $a_j^1 = \infty$.

By algebraic manipulation, we can show the following very important relationship:

$$\left(1 - F_{a_j}(s)\right) = \left(1 - F_{a_j^0}(s)\right)\left(1 - F_{a_j^1}(s)\right), \tag{36}$$

that is, $a_j = \min\{a_j^0, a_j^1\}$ is a *first order statistics* (David and Nagaraja, 2003). Thus, to simulate the inter-arrival time $a_j$, we need to sample only $a_j^0$ and $a_j^1$ and select the minimum. Simulating the inter-arrival time this way also provides us knowledge on the triggering source of the event time $t_j$. This is captured by the variable $X_j$ associated with each event time $t_j$, it takes the value of 0 when the event time $t_j$ is an immigrant, or 1 if $t_j$ is an offspring.

After sampling the inter-arrival time $a_j$, we compute the event time $t_j = t_{j-1} + a_j$ and sample the intensity jump size $Y_j$ to add to the intensity function. The intensity $\lambda^1(t)$ is cached for efficiency in simulation, its update rule follows

$$\lambda^1(t_j) = \lambda^1(t_{j-1})\,e^{-\delta(t_j - t_{j-1})} + Y_j. \tag{37}$$

This expression is similar to the recursive expression in Ozaki (1979), for which the current intensity value only depends on its previous value and the intensity jump size. In addition, we also update the counting process $N(t)$ by 1.

The above simulation method is repeated to sample each event time sequentially, and is terminated when the obtained event time exceeds the maturity time $T$. This last event time and its associated variables are discarded.

The sampling algorithm is summarised by Algorithm 2.

## Appendix B. Connections to Dassios and Zhao (2013)

An initial glance at Algorithm 1 may suggest that our simulation method is identical to that of Dassios and Zhao (2013) denoted by DZ here. We illustrate the differences between our work and DZ through the simulation of a bivariate Hawkes:

$$\lambda_1(t) = \mu_1 + Y_1^1(0)\,e^{-\delta_1^1 t} + Y_1^2(0)\,e^{-\delta_1^2 t} + \sum_{j=1: t \geq t_j^1}^{N^1(t)} Y_{1,j}^1\,e^{-\delta_1^1 t} + \sum_{j=1: t \geq t_j^2}^{N^2(t)} Y_{1,j}^2\,e^{-\delta_1^2 t}, \tag{38}$$

---

**Algorithm 2** Simulation of Univariate Hawkes

Given Input $\mu$, $Y_0$, $\delta$, $N_0$, $T$, and distribution for $Y_j$.

1. Initialise $t_0 = 0$ and let $\lambda^1(t_0) = Y_0$.

2. For $j = 1, 2, 3, \ldots$

   (a) Sample $a_j^0$ from the CDF in Equation (34).
   
   (b) Sample $a_j^1$ from the CDF in Equation (35).
   
   (c) Set $t_j = t_{j-1} + \min\{a_j^0, a_j^1\}$.
   
   (d) Set $X_j = \arg\min_i a_j^i$.
   
   (e) Sample $Y_j$ and update $\lambda^1(t_j)$ with Equation (37).
   
   (f) Update $N(t_j) = N(t_{j-1}) + 1$
   
   (g) Terminate algorithm if $t_j > T$, discard $t_j$ and its associated variables.

---

where $\lambda_1(t)$ is the intensity function for process 1. The intensity function $\lambda_2(t)$ is similar. The DZ equivalent, using our notation, would be (note the constant decays)

$$\lambda_1(t) = \mu_1 + Y_1^1(0) e^{-\delta_1^1 t} + \sum_{j=1: t \geq t_j^1}^{N^1(t)} Y_{1,j}^1 e^{-\delta_1^1 t} + \sum_{j=1: t \geq t_j^2}^{N^2(t)} Y_{1,j}^2 e^{-\delta_1^1 t}. \tag{39}$$

- **Different decay rates**. A major difference is that the decay rates $\delta_1^1$, $\delta_1^2$, $\delta_2^1$, and $\delta_2^2$ in our Hawkes model have different values, whereas the decays in DZ have to satisfy $\delta_1^1 = \delta_1^2$, and $\delta_2^1 = \delta_2^2$.

- **Multiple initial intensity jumps**. Our Hawkes process models multiple initial intensity jumps $Y_1^1(0)$, $Y_1^2(0)$, $Y_2^1(0)$, $Y_2^2(0)$, each with different decay rates, whereas in DZ there is only one such term in process 1, namely $\lambda_1(0) - a_1$ that is defined as $Y_1^1(0)$ in Equation (39). Similarly for process 2.

- **Markovian constraint.** In DZ, the requirement of constant decay rates in each process $m$ is needed for the ODE to be of the form: $\frac{d\lambda_m(t)}{dt} = -\delta_m^m(\lambda_m(t) - \mu_m)$. However, this nice property *breaks down* when the decay rates are dissimilar. In our case, we employ a different approach (superposition theory, see Section 4) to circumvent this constraint.

- **Origin of the event times**. The origin of an event time $r_j$ refers to the process that trigger its generation. With our method, we can easily decipher the origin of $r_j$ through the variable $X_j$. For example, choose an $r_j$ in process 1 ($Z_j = 1$), we know whether $r_j$ is an immigrant ($X_j = 0$), triggered by process 1 (self-excited, $X_j = 1$), or comes from process 2 (externally-excited, $X_j = 2$). Such interpretation does not seem possible with DZ.

- **Sampling different auxiliary variables.** Recall that the method of sampling simpler auxiliary variables (e.g., $a_j^i$) is employed to simulate the inter-arrival times. Although both DZ and our algorithm use this method, the specific auxiliary variables are *different*. To illustrate, DZ samples four variables for each event time $r_j$, associated with these four intensities: $\mu_1$, $\mu_2$, $\lambda_1(t) - \mu_1$, and $\lambda_2(t) - \mu_2$. In our case, however, we invoke the superposition theory to split the intensities to six pieces, allowing us determine the origin of the event times (See Section 4). For DZ: each inter-arrival time $r_j$ can be obtained by sampling four simpler random variables (this is possible because the decays are constant as mentioned above) and then the smallest is chosen, see details in Dassios and Zhao (2013, Algorithm 5.1). On the other hand, with our proposed method, the inter-arrival time $r_j$ is obtained by sampling six simpler random variables, namely, $a_{1j}^0$, $a_{1j}^1$, $a_{1j}^2$, $a_{2j}^0$, $a_{2j}^1$, and $a_{2j}^2$. This is an artifact from having dissimilar decays, at the same time, this method gives us an *interpretation* of branching type. For instance, the variables $a_{1j}^0$ and $a_{2j}^0$ correspond to an immigrant $r_j$; the variables $a_{1j}^1$ and $a_{2j}^2$ are attributed to self-excitation; while $a_{1j}^2$ and $a_{2j}^1$ are coming from external excitation.

- **Caching of intensity functions.** The treatment in caching the intensity functions is also different. DZ stores only the intensity functions $\lambda_1(t)$ and $\lambda_2(t)$ for simulation. In our algorithm, however, since the decays are dissimilar, we store the various components of the intensity functions, specifically, $\lambda_1^1(t)$, $\lambda_1^2(t)$, $\lambda_2^1(t)$, $\lambda_2^2(t)$. On updating the caches, while DZ can simply add $Y_m^i(s)$ into $\lambda_1(t)$ and $\lambda_2(t)$ (see Step 5 of Algorithm 5.1 in DZ), we need to add $Y_m^i(s)$ to the correct caches, see Equation (21).

We would like to point out that our simulation algorithm can be reduced to Dassios and Zhao (2013) when the decays in each process are constant, by doing the following. Again, we illustrate for the bivariate case. First, by redefining $Y_1^1(0) + Y_1^2(0)$ as $(\lambda_1(0) - a_1)$, $Y_2^1(0) + Y_2^2(0)$ as $(\lambda_2(0) - a_2)$, and setting $\delta_1^1 = \delta_1^2$ and $\delta_2^1 = \delta_2^2$, we obtain the same Hawkes model as that of Dassios and Zhao (2013). Second, we can use the superposition theory to group inter-arrival times from self-excitation and from external-excitation, that is, we group them into $\{a_{1j}^1, a_{1j}^2\}$ and $\{a_{2j}^1, a_{2j}^2\}$, while leaving $a_{1j}^0$ and $a_{2j}^0$ the same. Thirdly, this also means that we combine the intensities into $\lambda_1^1(t) + \lambda_1^2(t) = \lambda_1^*(t)$ and $\lambda_2^1(t) + \lambda_2^2(t) = \lambda_2^*(t)$. Note that we can still sample from the corresponding CDF without difficulty when we combine the intensities, this is because the decays in each process are now constant. Thus, we only need to sample four variables in this setup. Finally, since we have combined the intensities, the update rule for adding $Y$ is now the same as Dassios and Zhao (2013). To conclude, the simulation algorithm in Dassios and Zhao (2013) is a special case of our algorithm when the decay rates in each process are constant and after some simplification in our algorithm.

## Appendix C. Markov Chain Monte Carlo Method For Hawkes Processes

Here, we detail our proposed MCMC algorithm that utilises both adaptive rejection sampling (ARS, Gilks and Wild, 1992) and auxiliary variable augmentation (van Dyk and Meng, 2001). This algorithm is novel in that we propose a fully Gibbs sampler by using the ARS and show that the corresponding posteriors are log-concave. With this, we achieve no wastage (samples rejection) and thus all samples are useful.

### C.1. Branching Representation of Hawkes Processes

We note that the Hawkes processes can also be represented by a cluster process (Dassios and Zhao, 2013; Lee et al., 2016). For the MCMC algorithm, we employ this 'branching representation' which leads to a fully Gibbs sampler. This representation corresponds to introducing additional random variables called the auxiliary variables (see data augmentation, van Dyk and Meng, 2001). We note that in contrast to the representation described in Dassios and Zhao (2013) and Lee et al. (2016), we depict the branching representation for a multidimensional Hawkes process.

In this representation, we say an event time $t_j^m$ is an immigrant if it is generated from the background intensity $\mu_m$, otherwise, we say $t_j^m$ is an offspring. If $t_j^m$ is an offspring, it can be from the edge effect, or an offspring of another observed event time (either self excited or externally excited).

For each $t_j^m$, we introduce an indicator matrix $A_j^m$ that tells us the source of $t_j^m$. Collectively, all the $A_j^m$ tell us the branching structure of the Hawkes processes. We note that each $A_j^m$ is a special indicator matrix where only one of its element is 1. The particular entry of 1 tells us the type of the event time $t_j^m$:

1. If $t_j^m$ is an immigrant, $(A_j^m)_{00} = 1$.

2. If $t_j^m$ is an offspring due to edge effect from process $i$, then $(A_j^m)_{i0} = 1$.

3. If $t_j^m$ is an offspring of another event $t_k^i$ satisfying $t_k^i < t_j^m$, then $(A_j^m)_{ik} = 1$.

### C.2. Model Priors

We assume gamma priors for the parameters in the Hawkes process, noting that some of them are conjugate.

$$\mu_m \sim \text{Gamma}(\tau_{\mu_m}, \psi_{\mu_m}) \tag{40}$$

$$\delta_m^i \sim \text{Gamma}(\tau_{\delta_m^i}, \psi_{\delta_m^i}) \tag{41}$$

$$\alpha_m^i \sim \text{Gamma}(\tau_{\alpha_m^i}, \psi_{\alpha_m^i}) \tag{42}$$

$$\beta_m^i \sim \text{Gamma}(\tau_{\beta_m^i}, \psi_{\beta_m^i}) \tag{43}$$

where the $\tau_\circ > 0$, $\psi_\circ > 0$ for $\circ \in \Theta := \{\mu_m, \delta_m^i, \alpha_m^i, \beta_m^i\}$.

### C.3. Prior likelihood

For gamma distributed $\circ \in \{\mu_m, \delta_m^i, \alpha_m^i, \beta_m^i\}$, their priors are:

$$P(\circ \mid \tau_\circ, \psi_\circ) \propto \circ^{\tau_\circ - 1} e^{-\psi_\circ \circ} \tag{44}$$

More specifically:

$$P(\mu_m \mid \tau_{\mu_m}, \psi_{\mu_m}) \propto \mu_m^{\tau_{\mu_m} - 1} e^{-\psi_{\mu_m} \mu_m} \tag{45}$$

$$P(\delta_m^i \mid \tau_{\delta_m^i}, \psi_{\delta_m^i}) \propto {\delta_m^i}^{\tau_{\delta_m^i} - 1} e^{-\psi_{\delta_m^i} \delta_m^i} \tag{46}$$

$$P(\alpha_m^i \mid \tau_{\alpha_m^i}, \psi_{\alpha_m^i}) \propto {\alpha_m^i}^{\tau_{\alpha_m^i} - 1} e^{-\psi_{\alpha_m^i} \alpha_m^i} \tag{47}$$

$$P(\beta_m^i \mid \tau_{\beta_m^i}, \psi_{\beta_m^i}) \propto {\beta_m^i}^{\tau_{\beta_m^i} - 1} e^{-\psi_{\beta_m^i} \beta_m^i} \tag{48}$$

### C.4. Model Likelihood

Here, we use the branching representation of Hawkes processes described above. Note that *a priori* all branching structure is equally likely, thus for $\mathbf{A} = \{A_j^m\}$, we have

$$P(\mathbf{A}) = \prod_{m=1}^{M} \prod_{j=1}^{N^m(T)} P(A_j^m) \propto 1 \tag{49}$$

For the intensity jump sizes, the likelihood of $Y$ is

$$P(\mathbf{Y} \,|\, \alpha, \beta) = \prod_{m=1}^{M} \prod_{j=1}^{N^m(T)} \prod_{i=1}^{M} P(Y_{i,j}^m \,|\, \alpha_i^m, \beta_i^m)$$

$$= \prod_{m=1}^{M} \prod_{i=1}^{M} \prod_{j=1}^{N^i(T)} P(Y_{m,j}^i \,|\, \alpha_m^i, \beta_m^i)$$

$$= \prod_{m=1}^{M} \prod_{i=1}^{M} \prod_{j=1}^{N^i(T)} \frac{(\beta_m^i)^{\alpha_m^i}}{\Gamma(\alpha_m^i)} (Y_{m,j}^i)^{\alpha_m^i - 1} e^{-\beta_m^i Y_{m,j}^i} \tag{50}$$

The joint likelihood for all the event times is

$$P(\mathbf{t} \,|\, \mathbf{Y}, \boldsymbol{\Theta}, \mathbf{A}) = \prod_{m=1}^{M} \left( \prod_{j=1}^{N^m(T)} \lambda_m(t_j^m) \right) \exp\left( -\Lambda_m(T) \right) \tag{51}$$

where $\lambda_m(t_j^m)$ is the intensity that generated event time $t_j^m$:

$$\lambda_m(t_j^m) = (\mu_m)^{(A_j^m)_{00}} \prod_{i=1}^{M} \left[ \left( Y_m^i(0) \, e^{-\delta_m^i t_j^m} \right)^{(A_j^m)_{i0}} \prod_{k=1}^{N^i(T)} \left( Y_{m,k}^i \, e^{-\delta_m^i (t_j^m - t_k^i)} \right)^{(A_j^m)_{ik}} \right] \tag{52}$$

and $\Lambda_m(t) = \int_0^t \lambda_m(s) \, \mathrm{d}s$ is the compensator for process $m$, which can be computed as

$$\Lambda_m(t) = \mu_m t + \sum_{i=1}^{M} \left[ \frac{Y_m^i(0)}{\delta_m^i} \left( 1 - e^{-\delta_m^i t} \right) + \sum_{j=1}^{N^i(t)} \frac{Y_{m,j}^i}{\delta_m^i} \left( 1 - e^{-\delta_m^i (t - t_j^i)} \right) \right]. \tag{53}$$

The full joint likelihood for the Hawkes process can be written as follows:

$$P(\boldsymbol{\Theta} \,|\, \mathbf{t}, \mathbf{Y}) \propto P(\boldsymbol{\Theta}) \prod_{m=1}^{M} \left( \prod_{j=1}^{N^m(T)} \lambda_m(t_j^m) \right) e^{-\Lambda_m(T)} \prod_{i=1}^{M} \prod_{j=1}^{N^i(T)} (Y_{m,j}^i)^{\alpha_m^i - 1} e^{-\beta_m^i Y_{m,j}^i} \tag{54}$$

### C.5. Posterior Likelihoods and Gibbs Sampling

We derive Gibbs samplers for learning the parameters. For the variables $\mathbf{A}$, $\mathbf{Y}$, $\mu$, and $\beta$, we sample from their posteriors directly (since their posteriors follow known distributions), while for the parameters $\delta$, and $\alpha$, we adopt the adaptive rejection sampler to sample from their posteriors. In the following, we derive the posterior distributions for the parameters of interest.

### C.5.1. GIBBS SAMPLER FOR $A_j^m$

We first note that the posterior of $\mathbf{A}$ can be derived as

$$P(\mathbf{A} \,|\, \mathbf{t}, \mathbf{Y}, \boldsymbol{\Theta})$$
$$\propto \prod_{m=1}^{M} \prod_{j=1}^{N^m(T)} \left[ (\mu_m)^{(A_j^m)_{00}} \prod_{i=1}^{M} \left[ \left( Y_m^i(0)\, e^{-\delta_m^i t_j^m} \right)^{(A_j^m)_{i0}} \prod_{k=1}^{N^i(T)} \left( Y_{m,k}^i\, e^{-\delta_m^i (t_j^m - t_k^i)} \right)^{(A_j^m)_{ik}} \right] \right] \tag{55}$$

From this, we can see that each $A_j^m$ follows a Multinomial posterior:

$$A_j^m \,|\, \mathbf{t}, \mathbf{Y}, \boldsymbol{\Theta} \sim \text{Multinomial}(1, Q_j^m) \tag{56}$$

where $Q_j^m$ is a probability matrix, for $i = 1, \ldots, M$ and $k = 1, \ldots, N^i(T)$ that satisfy $t_j^m > t_k^i$.

$$(Q_j^m)_{00} = \mu_m / \lambda_m(t_j^m) \tag{57}$$

$$(Q_j^m)_{i0} = \left( Y_m^i(0)\, e^{-\delta_m^i t_j^m} \right) / \lambda_m(t_j^m) \tag{58}$$

$$(Q_j^m)_{ik} = \left( Y_{m,k}^i\, e^{-\delta_m^i (t_j^m - t_k^i)} \right) / \lambda_m(t_j^m) \tag{59}$$

In the Gibbs sampler, we sample new $A_j^m$ directly from its posterior.

### C.5.2. GIBBS SAMPLER FOR $Y_{m,j}^i$

For $j = 1, \ldots, N^i(T)$, and with a conjugate Gamma prior on $Y_{m,j}^i$, the posterior for $Y_{m,j}^i$ follows a Gamma distribution:

$$Y_{m,j}^i \,|\, \mathbf{t}, \mathbf{A}, \boldsymbol{\Theta} \sim \text{Gamma}\big((\alpha_m^i)^*, (\beta_m^i)^*\big) \tag{60}$$

where

$$(\alpha_m^i)^* = \alpha_m^i + \sum_{k=i}^{N^m(T)} (A_k^m)_{ij} \tag{61}$$

$$(\beta_m^i)^* = \beta_m^i + \frac{1}{\delta_m^i}\left(1 - e^{-\delta_m^i (T - t_k^i)}\right) \tag{62}$$

The posterior can easily be sampled.

The expression $\sum_{k=i}^{N^m(T)} (A_k^m)_{ij}$ represents the number of times $Y_{m,j}^i$ is 'used' by the subsequent event times. This value can be cached to improve algorithm performance.

### C.5.3. GIBBS SAMPLER FOR $Y_m^i(0)$

The Gibbs sampler for $Y_m^i(0)$ can be similarly derived.

$$Y_m^i(0) \,|\, \mathbf{A}, \boldsymbol{\Theta}, \tau, \psi \sim \text{Gamma}\left(\tau^*_{Y_m^i(0)}, \psi^*_{Y_m^i(0)}\right) \tag{63}$$

where

$$\tau^*_{Y^i_m(0)} = \tau_{Y^i_m(0)} + \sum_{j=1}^{N^m(T)} (A^m_j)_{i0} \tag{64}$$

$$\psi^*_{Y^i_m(0)} = \psi_{Y^i_m(0)} + \frac{1}{\delta^i_m}\left(1 - e^{-\delta^i_m T}\right) \tag{65}$$

### C.5.4. Gibbs Sampler for $\mu_m$

The posterior for $\mu_m$ can be derived easily:

$$\mu_m \mid \mathbf{A}, \mathbf{\Theta}, \tau, \psi \sim \text{Gamma}\left(\tau_{\mu_m} + \sum_{j=1}^{N^m(T)} (A^m_j)_{00}, \psi_{\mu_m} + T\right) \tag{66}$$

### C.5.5. Gibbs sampler for $\beta^i_m$

We note that Gamma priors on the $\beta$'s give Gamma posterior.

$$\beta^i_m \mid \mathbf{Y}, \tau, \psi \sim \text{Gamma}\left(\tau_{\beta^i_m} + N^i(T)\,\alpha^i_m \;,\; \psi_{\beta^i_m} + \sum_{i=1}^{N^i(T)} Y^i_{m,j}\right) \tag{67}$$

### C.5.6. Posterior for $\delta$

The posterior for $\delta^i_m$ can be derived as

$$P(\delta^i_m \mid \mathbf{t}, \mathbf{Y}, \mathbf{A}, \tau, \psi) \propto (\delta^i_m)^{\tau_{\delta^i_m} - 1} \exp\Bigg[-\delta^i_m\Bigg(\psi_{\delta^i_m} + \sum_{j=1}^{N^m(T)} \sum_{k=0:\, t^i_k < t^m_j}^{N^i(T)} (A^m_j)_{ik}(t^m_j - t^i_k)\Bigg)$$
$$- \sum_{k=0}^{N^i(T)} \frac{Y^i_{m,k}}{\delta^i_m}\left(1 - e^{-\delta^i_m(T - t^i_k)}\right)\Bigg] \tag{68}$$

where we have introduced $Y^i_{m,0}$ as a short hand for $Y^i_m(0)$ and $t_0 = 0$ to simplify the posterior. We note that this posterior is log-concave when certain conditions are met.

### C.5.7. Posterior for $\alpha$

The posterior for $\alpha^i_m$ can be derived as

$$P(\alpha^i_m \mid \mathbf{Y}, \beta, \tau, \psi) \propto (\alpha^i_m)^{\tau_{\alpha^i_m} - 1} [\Gamma(\alpha^i_m)]^{-N^i(T)} \left((\beta^i_m)^{N^i(T)} e^{-\psi_{\alpha^i_m}} \prod_{k=1}^{N^i(T)} Y^i_{m,k}\right)^{\alpha^i_m} \tag{69}$$

We note that this posterior is log-concave when certain conditions are met, this is detailed below. For ease of notation, we will denote the term inside the bracket in Equation (69) as $\kappa^i_m$, that is,

$$\kappa^i_m = (\beta^i_m)^{N^i(T)} e^{-\psi_{\alpha^i_m}} \prod_{k=1}^{N^i(T)} Y^i_{m,k}. \tag{70}$$

Note that this term is a *positive* constant with respect to $\alpha^i_m$.